\documentclass{article}

\usepackage{arxiv}
\usepackage{amsmath}
\usepackage[utf8]{inputenc} 
\usepackage[T1]{fontenc}    
\usepackage{hyperref}       
\usepackage{url}            
\usepackage{booktabs}       
\usepackage{amsfonts}       
\usepackage{nicefrac}       
\usepackage{microtype}      
\usepackage{lipsum}		
\usepackage{graphicx}
\usepackage[numbers,sort&compress]{natbib}
\usepackage{doi}
\usepackage{multirow}
\usepackage{bm}  
\usepackage{pifont}
\usepackage[final]{changes} 

\title{State of Health Estimation of Batteries Using a Time-Informed Dynamic Sequence-Inverted Transformer}

\author{{\hspace{1mm}Janak M. Patel}\thanks{Corresponding Author.}\\
	Applied Research, Quantiphi\\
        Marlborough, MA 01752, USA\\
	\texttt{janak.patel@quantiphi.com} \\
        \And
        {\hspace{1mm}Milad Ramezankhani}\\
        Applied Research, Quantiphi\\
        Marlborough, MA 01752, USA\\
	\texttt{milad.ramezankhani@quantiphi.com} \\
        \And
	{\hspace{1mm}Anirudh Deodhar} \\
	Applied Research, Quantiphi\\
        Marlborough, MA 01752, USA\\
	\texttt{anirudh.deodhar@quantiphi.com} \\
        \And
	{\hspace{1mm}Dagnachew Birru} \\
	Applied Research, Quantiphi\\
        Marlborough, MA 01752, USA\\
	\texttt{dagnachew.birru@quantiphi.com} \\
}

\renewcommand{\headeright}{}
\renewcommand{\shorttitle}{Accepted AI4TS Workshop at AAAI 2026}

\hypersetup{
pdftitle={A template for the arxiv style},
pdfsubject={q-bio.NC, q-bio.QM},
pdfauthor={David S.~Hippocampus, Elias D.~Striatum},
pdfkeywords={First keyword, Second keyword, More},
}

\begin{document}
\maketitle

\begin{abstract}
The rapid adoption of battery-powered vehicles and energy storage systems over the past decade has made battery health monitoring increasingly critical. Batteries play a central role in the efficiency and safety of these systems, yet they inevitably degrade over time due to repeated charge-discharge cycles. This degradation leads to reduced energy efficiency and potential overheating, posing significant safety concerns. Accurate estimation of a battery’s State of Health (SoH) is therefore essential for ensuring operational reliability and safety. Several machine learning architectures, such as LSTMs, transformers, and encoder-based models, have been proposed to estimate SoH from discharge cycle data. However, these models struggle with the irregularities inherent in real-world measurements: discharge readings are often recorded at non-uniform intervals, and the lengths of discharge cycles vary significantly. To address this, most existing approaches extract features from the sequences rather than processing them in full, which introduces information loss and compromises accuracy. To overcome these challenges, we propose a novel architecture: Time-Informed Dynamic Sequence Inverted Transformer (TIDSIT). TIDSIT incorporates continuous time embeddings to effectively represent irregularly sampled data and utilizes padded sequences with temporal attention mechanisms to manage variable-length inputs without discarding sequence information. Experimental results on the NASA battery degradation dataset show that TIDSIT significantly outperforms existing models, achieving over 50\% reduction in prediction error and maintaining an SoH prediction error below 0.58\%. Furthermore, the architecture is generalizable and holds promise for broader applications in health monitoring tasks involving irregular time-series data.
\end{abstract}


\section{Introduction}
The global transition toward clean energy and decarbonization is accelerating the adoption of electric vehicles (EVs) and large-scale battery energy storage systems. In response, automotive manufacturers are rapidly advancing electrification efforts, with several regions mandating that all new vehicles be battery-electric by 2035 to meet stringent emissions targets \citep{swarnkar2023systematic}. At the core of these technologies lie lithium-ion batteries, praised for their high energy density and efficiency \citep{lu2013review}. Despite their advantages, lithium-ion batteries naturally degrade over time as a result of repeated charge–discharge cycles. This degradation leads to diminished capacity, reduced efficiency, and heightened risks of thermal instability or failure \citep{belt2011calendar}. The consequences can be significant, ranging from safety hazards and unexpected system downtime to expensive battery replacements \citep{cabrera2016calculation}. Consequently, accurate estimation of a battery’s \textit{State of Health} (SoH) is critical for ensuring operational reliability, safety, and cost-effective lifecycle management \citep{venugopal2019state}.

In recent years, data-driven and machine learning (ML)-based approaches have gained significant traction for SoH estimation due to their ability to capture complex, nonlinear degradation patterns from historical data. Traditional SoH prediction methods typically rely on trends in discharge capacity to estimate the current life of a battery \citep{richardson2017gaussian, liu2014lithium}. However, discharge capacity alone may not sufficiently reflect internal degradation, particularly under diverse operational conditions. Consequently, more recent methods leverage multivariate time-series signals, such as voltage, current, and temperature profiles during discharge cycles,to enhance predictive accuracy and model robustness \citep{saha2007battery,saha2008uncertainty}. These richer input features enable the development of more adaptive and generalizable models across a wide range of usage scenarios. Deep learning architectures such as Long Short-Term Memory (LSTM) networks, Convolutional Neural Networks (CNNs), and Support Vector Regression (SVR) have been widely adopted for SoH estimation due to their ability to capture temporal dependencies and nonlinear degradation trends in battery data \citep{li2025machine}. Additionally, deep autoencoders have been employed to extract low-dimensional representations from high-dimensional cycle data while preserving degradation-related information \citep{xu2022physics}. Recently, transformer-based models—initially developed for natural language processing—have gained popularity in time-series applications, owing to expressiveness and scalability of their attention mechanisms \citep{vaswani2017attention}. In the context of battery state-of-health (SoH) prediction, a hybrid architecture combining convolutional neural networks (CNNs) with transformers has been proposed to enhance predictive performance \citep{gu2023novel}. A transformer-based model incorporating a dedicated feature extraction mechanism for SoH prediction has also been developed \citep{luo2023simple}. Furthermore, inverted Transformer (iTransformer) has been applied to battery SoH prediction, where variable-specific embeddings are used instead of conventional temporal encodings, enabling richer representation of multivariate data \citep{guirguis2024transformer}.

\begin{figure}[t]  
    \centering
    \includegraphics[width=0.6\textwidth]{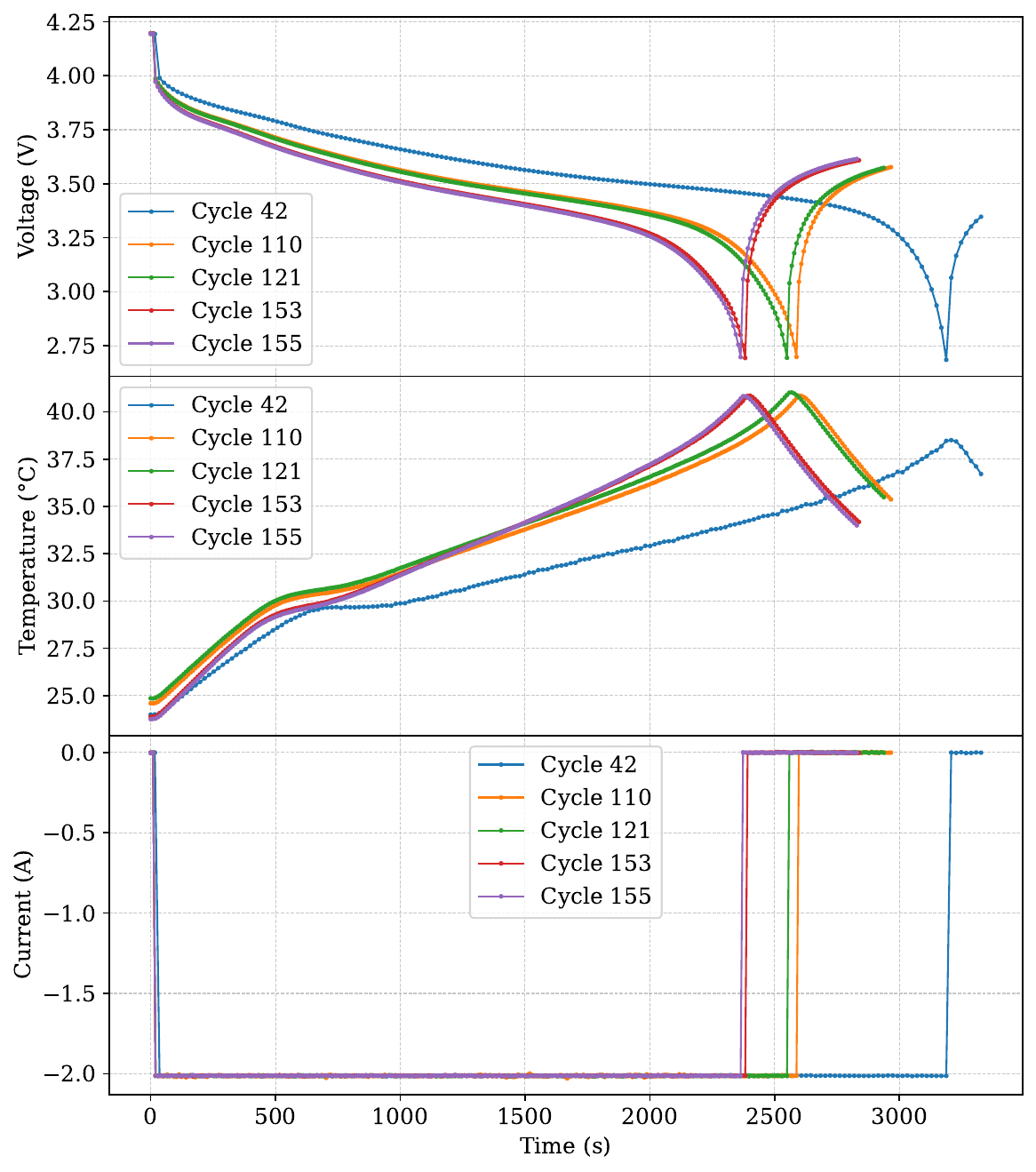} 
    \caption{Visualization of five randomly selected discharge cycles showing raw input features—voltage (V), temperature (°C), and current (A)—plotted against time. The figure highlights two key characteristics of real-world battery data: (1) \textit{irregular sampling intervals} within each cycle, and (2) \textit{variable sequence lengths} across cycles. Notably, later cycles become progressively shorter, illustrating the effect of battery degradation on discharge duration over time.}
    \label{fig:inputs}                             
\end{figure}

Despite the advancements in data-driven SoH estimation, existing models rely heavily on extensive feature engineering or sequence preprocessing to manage the practical challenges inherent in real-world battery data, such as \textit{irregular sampling intervals} and \textit{variable discharge cycle lengths} resulting from aging and inconsistent usage patterns as illustrated in Figure~\ref{fig:inputs}. To standardize these inputs, conventional methods typically employ interpolation or extract handcrafted statistical features from raw discharge sequences \citep{xu2022physics,guirguis2024transformer,li2025machine}. While such preprocessing simplifies the modeling pipeline, it often compromises temporal fidelity and may obscure subtle yet informative degradation patterns. Recent efforts have begun to move away from handcrafted features, leveraging deep learning models that can directly capture temporal dependencies. For instance, recurrent architectures like LSTMs have been used to process segments of discharge cycles while retaining sequential structure \citep{van2023estimation}. However, these models still depend on fixed-length input windows—typically derived from truncated or sampled portions of the full cycle—and therefore fail to exploit the complete discharge sequence in its raw, variable-length form. This limitation can result in the loss of long-range temporal dependencies and non-uniform degradation signals critical for accurate SoH estimation. These challenges underscore the need for a modeling framework capable of directly processing raw, irregularly sampled, and variable-length time series, while preserving the full temporal structure and improving both the robustness and generalizability of SoH prediction.

To overcome these critical limitations, we propose a novel deep learning architecture: the \textbf{Time-Informed Dynamic Sequence Inverted Transformer (TIDSIT)}. TIDSIT is designed from the ground up to directly address the challenges posed by \textit{irregularly sampled} and \textit{variable-length} time-series data in battery health monitoring, eliminating the need for lossy feature extraction or sequence truncation. Unlike conventional approaches that operate on preprocessed or fixed-length segments, TIDSIT processes entire discharge sequences in their native form—preserving complete temporal dynamics and enabling more faithful modeling of the battery's degradation trajectory. To effectively handle irregular sampling, TIDSIT incorporates \textit{continuous-time embeddings}, inspired by recent work in time-series forecasting \citep{kim2024continuous}. This embedding mechanism explicitly models the non-uniform intervals between observations, allowing the model to leverage temporal information. Additionally, TIDSIT incorporates data variate embeddings to explicitly model multivariate sensor inputs—such as voltage, current, and temperature—allowing the network to capture variable-specific temporal dynamics. This structured representation facilitates a more expressive encoding of battery behavior over time and is inspired by recent advances in multivariate time-series forecasting \citep{liu2023itransformer}. To accommodate variable-length discharge cycles, TIDSIT utilizes padded sequences along with a temporal attention mechanism that dynamically attends over all valid timestamps. By integrating these components within a transformer-based architecture, TIDSIT enables direct, end-to-end processing of raw discharge data, achieving accurate and generalizable SoH estimation without reliance on hand-engineered features. \textbf{The key contributions of this work are as follows:}
\begin{itemize}
    \item We present \textbf{TIDSIT}, a novel transformer-based architecture specifically designed for battery SoH estimation from raw discharge cycle data with irregular sampling intervals and variable sequence lengths.
    \item We address key limitations in existing SoH estimation approaches by introducing a suite of architectural innovations that enable robust learning from irregular, multivariate time series with variable sequence lengths. These include: (i) a \textbf{continuous-time embedding mechanism} to encode non-uniform temporal structures; (ii) \textbf{data variate embeddings} to capture feature-wise relationships among voltage, current, and temperature; and (iii) a \textbf{temporal attention mechanism} that operates on padded sequences to preserve full temporal resolution across variable-length discharge cycles.
\end{itemize}
\section{Methodology}
\subsection{Problem Specification}
The primary objective of this work is to develop a robust and generalizable deep learning model for estimating the \textit{SoH} of lithium-ion batteries, based on multivariate time-series data collected during individual discharge cycles. SoH is a key indicator of battery condition, commonly defined as the ratio between the current discharge capacity and the nominal (rated) capacity of the battery \citep{guo2014state}:

\begin{equation}
\text{SoH}(t) = \frac{C_{\text{current}}(t)}{C_{\text{rated}}}
\end{equation}

where $C_{\text{current}}(t)$ represents the discharge capacity measured at cycle $t$, and $C_{\text{rated}}$ denotes the capacity of a new, fully functional battery. The goal is to predict the SoH at each cycle using only the sensor signals recorded during that specific discharge cycle—without relying on information from future degradation cycles. Each discharge cycle is represented as a multivariate, irregularly sampled time series:

\begin{equation}
\mathcal{X}^{(i)} = \left\{ (x_1, \tau_1), (x_2, \tau_2), \ldots, (x_{T_i}, \tau_{T_i}) \right\}
\end{equation}

where $\mathcal{X}^{(i)}$ is the $i$-th discharge cycle, $x_j \in \mathbb{R}^d$ is a $d$-dimensional vector comprising sensor readings (e.g., voltage, current, temperature) recorded at time $\tau_j$, and $T_i$ is the number of observations in cycle $i$. Due to practical constraints in battery usage and asynchronous data logging, both the sequence length $T_i$ and the time intervals $\Delta \tau_j = \tau_j - \tau_{j-1}$ vary significantly across different cycles. The corresponding label for each cycle is the scalar SoH value:

\begin{equation}
y^{(i)} = \text{SoH}^{(i)} \in (0, 1]
\end{equation}

and the learning task is to identify a function $f_\theta$ parameterized by $\theta$, that maps each variable-length, irregularly sampled sequence to a scalar estimate of SoH:

\begin{equation}
\hat{y}^{(i)} = f_\theta(\mathcal{X}^{(i)})
\end{equation}

\noindent such that the prediction $\hat{y}^{(i)}$ closely approximates the ground truth label $y^{(i)}$ for all discharge cycles in the dataset.

\subsection{Time-Informed Dynamic Sequence Inverted Transformer (TIDSIT) Architecture}
\begin{figure}[htbp]  
    \centering
    \includegraphics[width=1\textwidth]{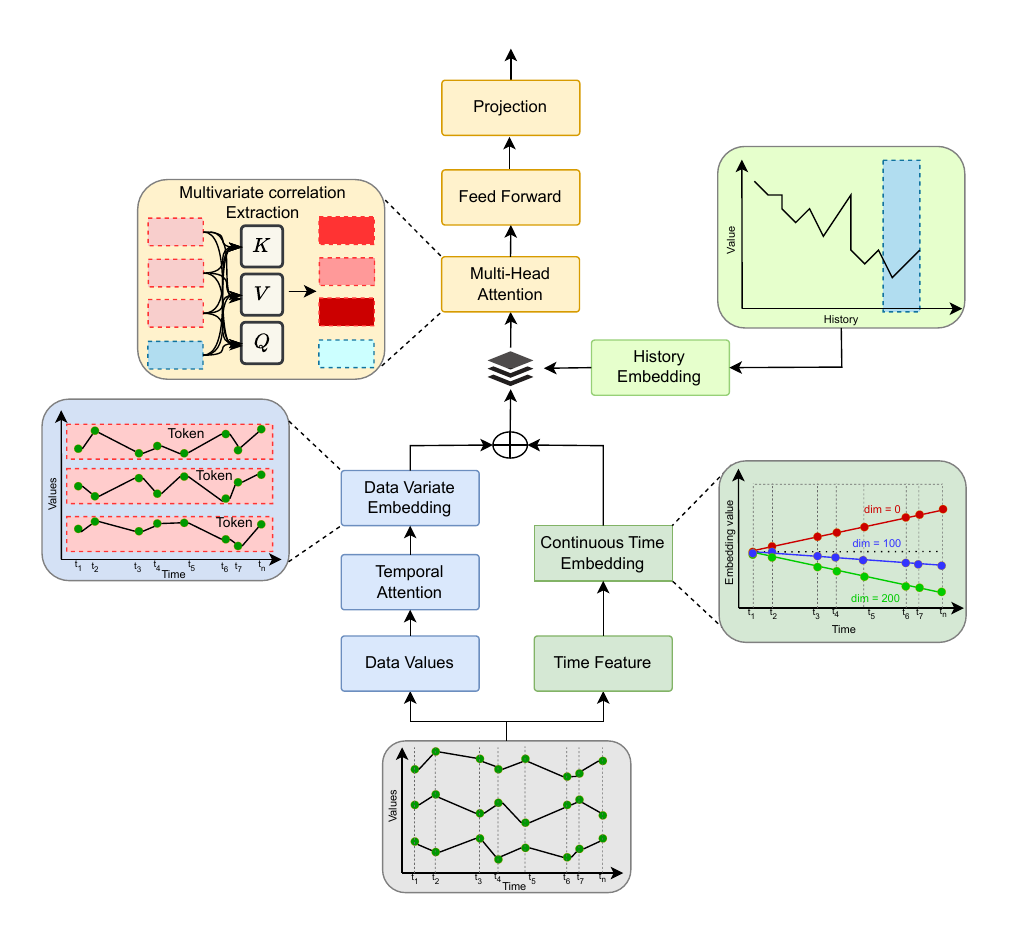} 
    \caption{Schematic of the proposed \textbf{Time-Informed Dynamic Sequence Inverted Transformer (TIDSIT)} architecture for battery SoH estimation. Each input consists of a multivariate, irregularly sampled discharge cycle, represented as a sequence of sensor readings—voltage, current, and temperature—along with corresponding timestamps. To accommodate varying sequence lengths, raw features are first passed through a \textit{temporal attention block} with padding masks, allowing the model to handle variable-length inputs without information loss. The output is processed by a \textit{data variate embedding} layer that learns a distinct representation for each sensor channel. In parallel, \textit{continuous-time embeddings} are computed based on the non-uniform timestamps and added to the variable-specific representations, preserving fine-grained temporal structure. A learned \textit{SoH history embedding}—derived from the SoH values of the past few discharge cycles—is then concatenated with the fused output. The resulting enriched sequence is passed through a transformer encoder where multi-head self-attention captures both intra-step interactions and long-range dependencies. Finally, the output is aggregated and passed through a feed-forward network and a projection head to yield the predicted scalar SoH for the current cycle.}
    \label{fig:system}                             
\end{figure}

To address the challenges posed by irregular sampling, variable-length inputs, and lack of historical context, we propose the \textbf{TIDSIT}. As illustrated in Figure~\ref{fig:system}, TIDSIT is a transformer-based architecture that processes raw multivariate discharge sequences—comprising asynchronous voltage, current, and temperature readings—without the need for truncation, interpolation, or handcrafted features.

TIDSIT is built upon four key components: (1) a \textit{continuous-time embedding} that encodes irregular timestamp information; (2) a \textit{data variate embedding module} that learns sensor-wise latent representations; (3) a \textit{temporal attention block} that mitigates the effects of sequence padding while capturing temporal dependencies within features; and (4) a \textit{SoH history embedding} that incorporates contextual trends from prior cycles. These representations are fused and passed through a transformer encoder, which models both intra-cycle sensor dynamics and inter-cycle degradation trends to generate accurate SoH estimates across diverse batteries and usage profiles.

\paragraph{Temporal Attention.}
To effectively process sequences of varying lengths, we pad each input to a fixed length \( T \), where \( T \) denotes the maximum sequence length observed across the dataset. Let each original discharge cycle be represented as \( X_o^{(i)} = [x_1, x_2, \ldots, x_{T_i}] \in \mathbb{R}^{T_i \times d} \), where \( T_i \) is the number of vthe number of observations of \( i \)-th cycle, and \( d \) is the number of sensor features (e.g., voltage, current, temperature). Sequences shorter than \( T \) are end-padded with a sentinel value of \(-1\), resulting in inputs of shape \( \mathbb{R}^{T \times d} \). Since all sensor readings are normalized to the range \([0, 1]\), this padding value is clearly distinguishable and does not conflict with valid measurements. To prevent these artificial values from influencing the learning process, we introduce a \textit{temporal attention block} that applies a binary padding mask. This mask ensures that attention scores involving padded positions are excluded from computation, allowing subsequent layers to ignore non-informative inputs. Additionally, this block captures temporal dependencies across the valid (unpadded) portions of each sequence before any data variate embedding is applied, enabling early reasoning about discharge dynamics. Given the padded input sequence \( X^{(i)} = [x_1, x_2, \ldots, x_T] \in \mathbb{R}^{T \times d} \), the scaled dot-product attention mechanism is computed as:

\begin{equation}
\text{Attention}(Q, K, V) = \text{softmax}\left( \frac{QK^\top}{\sqrt{d_k}} + M \right)V
\label{eq:temporal_attention}
\end{equation}

where $Q = XW_Q$, $K = XW_K$, $V = XW_V$, and $W_Q, W_K, W_V \in \mathbb{R}^{d \times d_k}$ are learnable projection matrices. The mask matrix $M \in \mathbb{R}^{T \times T}$ assigns large negative values to padded positions to eliminate their influence during attention computation \citep{devlin2019bert}. Multiple attention heads are combined as:

\begin{equation}
\text{MultiHead}(X) = \text{Concat}(\text{head}_1, \ldots, \text{head}_h)W_O
\end{equation}

where \( W_O \in \mathbb{R}^{(h \cdot d_k) \times d} \) is a learnable output projection matrix that maps the concatenated output of all attention heads back to the original model dimension \( d \). This is followed by a residual connection and layer normalization to stabilize training and improve gradient flow \citep{vaswani2017attention}:

\begin{equation}
Z = \text{LayerNorm}(X + \text{Dropout}(\text{MultiHead}(X)))
\end{equation}

The final output of this block, denoted $Z^{(i)} \in \mathbb{R}^{T \times d}$, is a temporally attended sequence retaining the original time step structure and feature dimensionality. This serves as the input to the data variate embedding module.
\paragraph{Data Variate Embedding.}
To capture inter-variable correlations across the entire discharge cycle, we implement a data variate embedding strategy that operates on the output of the temporal attention block \citep{liu2023itransformer}. Let the attended sequence output from the previous block be denoted as $Z^{(i)} \in \mathbb{R}^{T \times d}$. We transpose this to obtain $Z^{(i)}_{\text{var}} \in \mathbb{R}^{d \times T}$, where each row $z^{(v)}_{1:T} \in \mathbb{R}^{T}$ corresponds to the full time series of the $v$-th variable. Each variable sequence is then independently projected into a shared latent space:
\begin{equation}
z^{(v)} = z^{(v)}_{1:T}W^{(v)} + b^{(v)}, \quad \text{for } v = 1, \ldots, d.
\end{equation}
where $W^{(v)} \in \mathbb{R}^{T \times h}$ and $b^{(v)} \in \mathbb{R}^h$ are learnable parameters. The set of embeddings $\{z^{(1)}, \ldots, z^{(d)}\}$ is stacked as:
\begin{equation}
Z^{\text{var}} = [z^{(1)}; z^{(2)}; \ldots; z^{(d)}] \in \mathbb{R}^{d \times h}.
\end{equation}
enabling the model to learn high-level representations for each sensor signal that preserve its unique temporal behavior.

\paragraph{Continuous-Time Embedding.}
To incorporate fine-grained temporal structure from irregularly sampled sequences, we apply a continuous-time embedding scheme inspired by \citep{kim2024continuous}. Given the timestamp vector $\boldsymbol{\tau}^{(i)} = [\tau_1, \tau_2, \ldots, \tau_{T}]$, we normalize it to $[0, 1]$. The normalized vector is projected:
\begin{equation}
E^{\text{time}} = \boldsymbol{\tau} W_{\text{time}} + b_{\text{time}}, \quad E^{\text{time}} \in \mathbb{R}^{1 \times h}.
\end{equation}
with $W_{\text{time}} \in \mathbb{R}^{T \times h}$ and $b_{\text{time}} \in \mathbb{R}^h$. To retain information about the temporal ordering and relative timing of events within each irregularly sampled sequence—analogous to positional encoding in large language models—we add the continuous-time embedding to the data variate embedding:
\begin{equation}
Z^{\text{fused}} = Z^{\text{var}} + E^{\text{time}}
\end{equation}

\paragraph{SoH History Embedding.}
To encode information from previous discharge cycles, we maintain a fixed-length vector of past SoH values $[y^{(i-1)}, \ldots, y^{(i-p)}] \in \mathbb{R}^p$, where $p$ is the history window size. This sequence is mapped via a learned projection:
\begin{equation}
E^{\text{hist}} = [y^{(i-1)}, \ldots, y^{(i-p)}] W_{\text{hist}} + b_{\text{hist}}, \quad E^{\text{hist}} \in \mathbb{R}^{1 \times h}.
\end{equation}
where $W_{\text{hist}} \in \mathbb{R}^{p \times h}$ and $b_{\text{hist}} \in \mathbb{R}^h$ are learnable. The final sequence representation passed to the transformer encoder is formed by concatenating the fused embedding $Z^{\text{fused}}$ with the SoH history embedding $E^{\text{hist}}$ along the temporal dimension:
\begin{equation}
Z^{\text{input}} = [Z^{\text{fused}}; E^{\text{hist}}] \in \mathbb{R}^{(d+1) \times h}.
\end{equation}
allowing the encoder to model intra-cycle sensor dynamics, temporal trends, and inter-cycle degradation history jointly.

\paragraph{Transformer Encoder and Output.}
The final input $Z^{\text{input}}$ is processed by a Transformer encoder consisting of a multi-head self-attention layer followed by a position-wise feed-forward network. The encoded representation is then passed through a projection layer to yield the final SoH prediction $\hat{y}^{(i)}$.

\section{Results and Discussion}
\subsection{Datasets and Benchmark}
\subsubsection{Dataset Description}
We evaluate the proposed TIDSIT architecture using the battery degradation dataset provided by the NASA Ames Prognostics Center of Excellence (PCoE) \citep{saha2007battery}. This dataset consists of sensor measurements collected from lithium-ion 18650 battery cells undergoing repeated charge-discharge cycles under controlled aging protocols and varying ambient temperatures. For each discharge cycle, the dataset provides multivariate time-series data including voltage, current, and temperature readings. These three sensor channels—voltage (V), current (I), and temperature (T)—serve as the input features to our model. The target output is the battery's remaining capacity, measured at the end of each discharge cycle, which is used to compute the ground-truth SoH. During the aging process, periodic capacity tests were conducted to track the progressive degradation of each battery. The end-of-life (EOL) criterion is defined as a 30\% reduction in capacity from the nominal rating (i.e., from 2 Ah to 1.4 Ah). Consequently, each cycle is represented as a variable-length multivariate sequence of [V, I, T] signals along with a single scalar SoH label.

To assess the model's generalization capability, we follow a cross-battery evaluation setup. Specifically, data from batteries B0005 and B0006 are used for training, while battery B0007 is reserved for testing. This ensures that the model is evaluated on entirely unseen degradation patterns, simulating a realistic deployment scenario. This dataset presents a challenging setting for data-driven SoH estimation, characterized by irregular sampling intervals, varying sequence lengths, and heterogeneous degradation trajectories—making it well-suited for benchmarking the robustness of the proposed TIDSIT architecture.
\subsubsection{Benchmark Models}
To assess the performance of the proposed TIDSIT architecture, we compare it against a range of baseline models that represent both conventional and recent approaches to SoH estimation:

\begin{itemize}
    \item \textbf{Feedforward Neural Network (FNN):} A fully connected network trained on fixed-length subsequences extracted from the discharge data, rather than processing the entire variable-length sequence \citep{van2023estimation}.
    
    \item \textbf{LSTM (Fixed-range):} A Long Short-Term Memory network trained on fixed-length subsequences extracted from the discharge data, rather than processing the entire variable-length sequence \citep{van2023estimation}.
    
    \item \textbf{LSTM (Feature Extraction):} A feature-driven approach in which handcrafted features (e.g., voltage ranges, cycle durations) are extracted from each cycle and used to train an LSTM model for SoH prediction \citep{li2025machine}.
    
    \item \textbf{i-Transformer:} A recent transformer-based model that employs an inverted embedding scheme and relies on extracted features rather than directly processing raw time-series data \citep{guirguis2024transformer}.
\end{itemize}

These baselines provide a comprehensive comparison framework, enabling us to evaluate TIDSIT’s ability to perform end-to-end learning directly from raw, irregularly sampled sequences, in contrast to models that depend on windowed inputs or manual feature engineering.

\subsection{Model Hyperparameters}
The hyperparameters of the TIDSIT model were selected based on preliminary experiments and established design practices for transformer-based time series modeling. We set the model's hidden dimension to 42, with a single encoder layer and 8 attention heads to balance representational capacity and computational efficiency. A dropout rate of 0.1 is used for regularization, and the feed-forward network dimension is set to 168 (four times the hidden dimension), following standard transformer design principles. ReLU is used as the activation function. To accommodate variable-length discharge cycles within a batch, we pad all sequences to a maximum length of 371—corresponding to the longest sequence in the dataset—and use masking to ignore padded positions during attention computation. The model is trained on these padded sequences with a past SoH history window of size 10 provided as additional input context. All experiments were conducted on a standard laptop equipped with an 11th Gen Intel(R) Core(TM) i5-1135G7 CPU running at 2.40GHz and 8.00 GB RAM. No GPU acceleration was used during training or evaluation. Despite the limited computational resources, the proposed TIDSIT model achieved efficient training times, underscoring its practicality for lightweight deployment.

\subsection{Results}
To evaluate the effectiveness of the proposed TIDSIT architecture, we assess its performance on the B0007 battery dataset, which was entirely held out during training. Table~\ref{tab:my-table} reports the SoH estimation accuracy of TIDSIT compared to several benchmark models, including FNN \citep{van2023estimation}, LSTM (fixed-range \citep{van2023estimation} and feature-based variants \citep{li2025machine}), and the i-Transformer \citep{guirguis2024transformer}. TIDSIT achieves an RMSE of 0.0047, which marks a significant improvement over all baselines. Notably, it outperforms the best baseline—LSTM (Fixed-range)—which records an RMSE of 0.0082. This corresponds to more than a \textbf{50\% reduction in prediction error}, highlighting TIDSIT's superior ability to learn directly from raw, irregularly sampled time series. The FNN baseline performs worst, with an RMSE of 0.032, underscoring the limitations of static input modeling. In terms of RMSE percentage, TIDSIT achieves 0.58\%, outperforming the i-Transformer's 1.47\%, despite the latter relying on extracted statistical features. This demonstrates TIDSIT’s ability to generalize effectively without handcrafted features or preprocessing.

\begin{figure}[t]  
    \centering
    \includegraphics[width=1\textwidth]{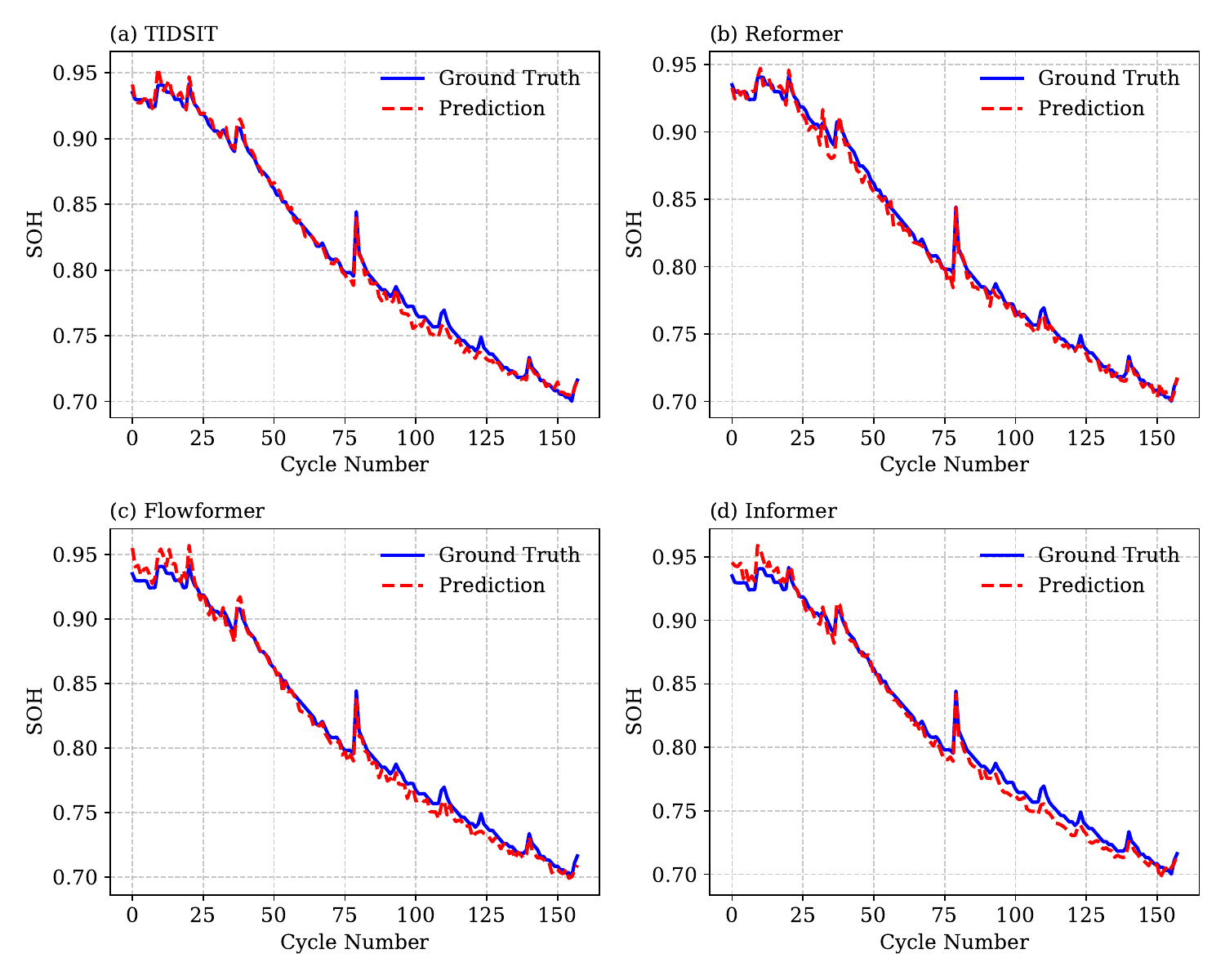} 
    \caption{Predicted versus ground truth SoH values for the B0007 battery (unseen during training). (a)–(d) compare the SoH prediction performance of different transformer variants: (a) TIDSIT, (b) Reformer, (c) Flowformer, and (d) Informer. The figure demonstrates the model’s generalization capability by accurately tracking the degradation trend on unseen test data.TIDSIT demonstrates the best alignment with the true degradation trajectory, indicating superior generalization to real-world battery behavior with irregular sampling and variable cycle lengths.}
    \label{fig:vali}                             
\end{figure}

Figure~\ref{fig:vali} illustrates the predicted versus ground truth SoH values for the B0007 battery. The predictions closely follow the true degradation trajectory, confirming TIDSIT’s capacity to generalize to unseen battery instances and accurately capture both long-term trends and short-term variations.

\begin{table}[]
\resizebox{\columnwidth}{!}{%
\begin{tabular}{@{}c|c|c|c|c|c@{}}
\toprule
Metric       & FNN   & LSTM (Fixed-range) & LSTM (Feature Extraction) & i-Transformer & TIDSIT \\ \midrule
RMSE         & 0.032 & 0.0082             & 0.0094                    & -             & 0.0047 \\ \midrule
RMSE Percent & -     & -                  & -                         & 1.47          & 0.58   \\ \bottomrule
\end{tabular}%
}
\caption{Comparison of SoH estimation performance across benchmark models using Root Mean Square Error (RMSE) and RMSE percentage on the B0007 test battery dataset.}
\label{tab:my-table}
\end{table}

\subsection{Comparison with Transformer Variants}
To isolate the contribution of the encoder architecture in SoH estimation, we perform a comparative analysis by replacing the Transformer encoder block in TIDSIT with several well-known Transformer variants designed for long-sequence modeling: \textbf{Reformer} \citep{kitaev2020reformer}, \textbf{Informer} \citep{zhou2021informer}, and \textbf{Flowformer} \citep{huang2022flowformer}. Importantly, all other components of the TIDSIT pipeline—such as the temporal attention block, data variate embedding, continuous-time embedding, and SoH history embedding—are retained unchanged across all variants. This design choice ensures that the comparison focuses solely on the encoder’s ability to capture intra-cycle and inter-cycle dependencies. Table~\ref{tab:encoder_variants} summarizes the SoH prediction performance of these encoder variants on the B0007 test battery. While Reformer, Informer, and Flowformer show competitive results, TIDSIT outperforms them all, achieving the lowest RMSE of 0.0047 and the lowest RMSE percentage of 0.58\%.
\begin{table}[h] 
\centering
\begin{tabular}{@{}c|c|c|c|c@{}}
\toprule
\textbf{Metric} & \textbf{Reformer} & \textbf{Informer} & \textbf{Flowformer} & \textbf{TIDSIT} \\ \midrule
RMSE           & 0.005             & 0.0071            & 0.0065              & \textbf{0.0047} \\ \midrule
RMSE Percent   & 0.61              & 0.89              & 0.80                & \textbf{0.58}   \\ \bottomrule
\end{tabular}%
\caption{Comparison of SoH estimation performance using different Transformer encoder variants. Only the encoder block is modified; all other components of TIDSIT are retained.}
\label{tab:encoder_variants}
\end{table}

\subsection{Ablation Study}
To evaluate the contribution of each architectural component in TIDSIT, we conduct an ablation study by systematically removing one module at a time and measuring the resulting SoH prediction performance. The modules evaluated include: (1) the \textit{Temporal Attention Block}, which handles variable-length sequences and captures time-step dependencies; (2) the \textit{Continuous-Time Embedding}, which encodes irregular sampling intervals; (3) the \textit{Data Variate Embedding}, which enables variable-wise representation learning; and (4) the \textit{SoH History Embedding}, which incorporates contextual information from previous cycles.

Table~\ref{tab:ablation} reports the RMSE and RMSE percentage on the unseen B0007 test battery for each variant. The full TIDSIT model achieves the best performance, with an RMSE of 0.0047 and RMSE percentage of 0.58\%. Removing any single component leads to a noticeable drop in performance, highlighting the importance of each module. Among all components, removing the \textit{Continuous-Time Embedding} causes the most significant degradation (RMSE = 0.0369), underscoring its critical role in modeling temporal irregularity inherent in real-world battery data. The absence of the \textit{Temporal Attention Block} also results in a substantial error increase (RMSE = 0.0190), reflecting its importance in handling padded inputs and capturing intra-cycle dependencies early in the pipeline. Excluding the \textit{Data Variate Embedding} leads to a moderate increase in error (RMSE = 0.0058), but also introduces a large computational overhead—training time increases from 40.4 minutes (with variate embedding) to 154.4 minutes (without it). This is because, without variate embedding, the model must process each time step as an independent token, significantly increasing the sequence length and computational cost. This highlights the dual benefit of the variate embedding: improved performance and substantially faster training. Interestingly, the removal of the \textit{SoH History Embedding} causes only a minor performance drop (RMSE = 0.0065), suggesting that while historical degradation trends provide complementary context, the model predominantly learns to infer SoH from the current discharge sequence itself. This indicates that TIDSIT is not simply correlating past SoH values but is effectively learning degradation dynamics directly from raw sensor inputs.
\begin{table}[h]
\centering
\begin{tabular}{@{}c|c|c|c|c|c|c@{}}
\toprule
\textbf{Configuration} &
  \textbf{\begin{tabular}[c]{@{}c@{}}History \\ Embedding\end{tabular}} &
  \textbf{\begin{tabular}[c]{@{}c@{}}Temporal \\ Attention\end{tabular}} &
  \textbf{\begin{tabular}[c]{@{}c@{}}Continuous-Time \\ Embedding\end{tabular}} &
  \textbf{\begin{tabular}[c]{@{}c@{}}Variate \\ Embedding\end{tabular}} &
  \textbf{RMSE} &
  \textbf{RMSE \%} \\ \midrule
\begin{tabular}[c]{@{}c@{}}Full Model \\ (TIDSIT)\end{tabular} &
  \checkmark &
  \checkmark &
  \checkmark &
  \checkmark &
  \textbf{0.0047} &
  \textbf{0.58} \\ \midrule
\begin{tabular}[c]{@{}c@{}}w/o Variate \\ Embedding\end{tabular} &
  \checkmark &
  \checkmark &
  \checkmark &
  \texttimes &
  0.0058 &
  0.71 \\ \midrule
\begin{tabular}[c]{@{}c@{}}w/o History \\ Embedding\end{tabular} &
  \texttimes &
  \checkmark &
  \checkmark &
  \checkmark &
  0.0065 &
  0.81 \\ \midrule
\begin{tabular}[c]{@{}c@{}}w/o Temporal \\ Attention\end{tabular} &
  \checkmark &
  \texttimes  &
  \checkmark &
  \checkmark &
  0.0190 &
  2.52 \\ \midrule
\begin{tabular}[c]{@{}c@{}}w/o Continuous Time \\ Embedding\end{tabular} &
  \checkmark &
  \checkmark &
  \texttimes  &
  \checkmark &
  0.0369 &
  4.79 \\ \bottomrule
\end{tabular}
\caption{Ablation study showing the impact of removing each architectural component from the full TIDSIT model on SoH estimation performance. 
"w/o" indicates that the corresponding component is removed from the full model configuration.}
\label{tab:ablation}
\end{table}

\section{Conclusion}
In this work, we introduced the \textbf{Time-Informed Dynamic Sequence Inverted Transformer (TIDSIT)}, a novel transformer-based architecture for accurate and generalizable estimation of battery State of Health (SoH) using raw multivariate discharge sequences. By effectively handling irregular sampling, variable-length sequences, and limited degradation history, TIDSIT enables end-to-end learning without the need for handcrafted feature extraction or sequence truncation. Our experiments demonstrate that TIDSIT significantly outperforms traditional models such as feedforward neural networks and LSTMs, as well as recent transformer variants, achieving more than 50\% reduction in prediction error on unseen test data. Furthermore, ablation studies highlight the critical importance of each architectural component—including temporal attention, continuous-time embedding, data variate embedding, and SoH history context—in achieving robust SoH predictions. Overall, TIDSIT establishes a strong foundation for reliable battery health monitoring and paves the way for future research into scalable, interpretable, and real-time prognostics using deep sequence models.


\appendix






\end{document}